\documentclass[sigconf,nonacm]{acmart}
\usepackage{multirow}
\usepackage{colortbl}
\usepackage{float}
\usepackage{subcaption}
\AtBeginDocument{%
  \providecommand\BibTeX{{%
    \normalfont B\kern-0.5em{\scshape i\kern-0.25em b}\kern-0.8em\TeX}}}

\begin{document}

\title{MoFE-Time:  Mixture of Frequency Domain Experts for Time-Series Forecasting models}


\author{Yiwen Liu}
\affiliation{%
  \institution{Li Auto Inc.}
  \city{Beijing}
  \country{China}}
\email{liuyiwen@lixiang.com}

\author{Chenyu Zhang}
\affiliation{%
  \institution{Li Auto Inc.}
  \city{Beijing}
  \country{China}}
\email{zhangchenyu@lixiang.com}

\author{Junjie Song}
\affiliation{%
  \institution{Li Auto Inc.}
  \city{Beijing}
  \country{China}}
\email{songjunjie@lixiang.com}

\author{Siqi Chen}
\affiliation{%
  \institution{Li Auto Inc.}
  \city{Beijing}
  \country{China}}
\email{chensiqi5@lixiang.com}

\author{Yin Sun}
\affiliation{%
  \institution{Li Auto Inc.}
  \city{Beijing}
  \country{China}}
\email{sunyin@lixiang.com}

\author{Zihan Wang}
\affiliation{%
  \institution{School of Artificial Intelligence , UCAS}
  \city{Beijing}
  \country{China}}
\email{wangzihan242@mails.ucas.ac.cn}

\author{Lingming Zeng}
\affiliation{%
  \institution{Li Auto Inc.}
  \city{Beijing}
  \country{China}}
\email{zenglingming@lixiang.com}

\author{Yuji Cao*}
\affiliation{%
  \institution{Li Auto Inc.}
  \city{Beijing}
  \country{China}}
\email{caoyuji@lixiang.com}

\author{Junming Jiao*}
\affiliation{%
  \institution{Li Auto Inc.}
  \city{Beijing}
  \country{China}}
\email{jiaojunming@lixiang.com}



\begin{abstract}
  As a prominent data modality task, time series forecasting plays a pivotal role in diverse applications. With the remarkable advancements in Large Language Models (LLMs), the adoption of LLMs as the foundational architecture for time series modeling has gained significant attention. Although existing models achieve some success, they rarely both model time and frequency characteristics in a pretraining-finetuning paradigm leading to suboptimal performance in predictions of complex time series, which requires both modeling periodicity and prior pattern knowledge of signals. 
 We propose MoFE-Time, an innovative time series forecasting model that integrates time and frequency domain features within a Mixture of Experts (MoE) network. Moreover, we use the pretraining-finetuning paradigm as our training framework to effectively transfer prior pattern knowledge across pretraining and finetuning datasets with different periodicity distributions.
 Our method introduces both frequency and time cells as experts after attention modules and leverages the MoE routing mechanism to construct multidimensional sparse representations of input signals. 
In experiments on six public benchmarks, MoFE-Time has achieved new state-of-the-art performance, reducing MSE and MAE by 6.95\% and 6.02\% compared to the representative methods Time-MoE.
 Beyond the existing evaluation benchmarks, we have developed a proprietary dataset, NEV-sales, derived from real-world business scenarios.  Our method achieves outstanding results on this dataset, underscoring the effectiveness of the MoFE-Time model in practical commercial applications.
 
 
\end{abstract}





\maketitle

\section{Introduction}
Time series data serves as a fundamental modality within dynamic real-world systems \cite{box2015time,zhang2024self,liang2024foundation}. As a critical task with significant demand across numerous fields—including  sales, energy, climate science, quantitative finance, and urban computing \cite{jin2023large,nie2024survey,mao2024time}, the prediction of time series has been the subject of extensive research over the years.


With the success of LLMs in the field of natural language processing, there is increasing interest in adopting large language model training methods for time series forecasting. Approaches like TimeCMA\cite{liu2024timecma,llm4s} attempt to directly integrate pre-trained LLMs with time series data for modality fusion, utilizing the language modal pretrain knowledge from LLMs to enhance time series forecasting performance. 
The alignment challenge between the language space of LLMs and the feature space of time series data is a significant obstacle for these methods.
Therefore, some methods attempt to train foundation models directly on time series modes. Time-MoE\cite{shi2024time} scaled a time series foundation model up to 2.4 billion parameters, achieving significantly improved forecasting precision has aroused widespread concern. However, these methods ignore the complex periodicity and non-stationarity inherent in time series modal.

It is well-established that frequency domain features are considered direct representations of signal periodicity. Currently, many deep learning-based approaches, such as FEDformer\cite{zhou2022fedformer}, leverage frequency domain features of signals to enhance model feature comprehension. However, these methods predominantly convert input signals directly into the frequency domain for learning, thereby introducing additional mathematical constraints and lacking the ability to effectively learn signal frequency domain representations, which results in suboptimal performance and inadequate generalization.

In response to the aforementioned issues, we aim to construct a time series forecasting network capable of directly learning the Fourier transform process that converts signals from the time domain to the frequency domain, thereby capturing the inherent properties of signals.  With the success of Deepseek-v3\cite{liu2024deepseek} in the large language model domain, networks based on the MOE architecture are gaining attention due to their inherent sparse attributes, which align naturally with our goal of modeling frequency domain features of signals. 
In addition, referring to the previously discussed methods, which enhance the capability of time series forecasting models through the integration of pre-trained large language models, we posit that directly applying a pretraining-finetuning approach to time series models can endow the models with prior knowledge of signals, thereby improving their performance and generalization ability.

We introduce MoFE-Time, a time series forecasting model based on the MoE architecture, which integrate  both frequency and time features into each expert module to learn both the temporal features and frequency domain intrinsic properties of signals.
By introducing reversible instance normalisation (RevIN)\cite{kim2021reversible} and time aggregation methods, the ability of the model to deal with non-stationarity and the applicability of the model to deal with variable-length series are improved.   Frequency-Time Cell (FTC) for Domain Experts enhancing the model's ability to capture the intrinsic properties of the signal's time and frequency domain, which contributes to improved predictive performance of the model.
In experiments on six public benchmarks, MoFE-Time has achieved new  \textbf{state-of-the-art} performance, reducing MSE and MAE by  \textbf{6.95\%} and \textbf{6.02\%} compared to the representative methods Time-MoE.


In summary, the main contributions of our paper are:
\begin{itemize}
\item 
We introduce MoFE-Time, a time series forecasting model based on the MoE architecture, which integrate  both frequency and time features into each expert module to learn both the temporal features and frequency domain intrinsic properties of signals. 

\item We adopt the pretraining-finetuning paradigm for joint time-frequency learning, and improve the performance and generalization of the model by introducing prior knowledge.

\item We collected NEV-sales, a dataset contains daily customer count of about 500 sales centers of a leading new energy vehicle brand across the country from 2022 to 2025. 

\item  Experimental results demonstrate that our method achieve SOTA on six public datasets and NEV-sales, which demonstrates the effectiveness of the MoFE-Time model.

\end{itemize}

\section{Related Work}

\textbf{Time Series Forecasting}.
In recent years, deep learning-based forecasting methods have been widely applied in time-series forecasting tasks.
Initially, recurrent neural networks (RNN) and their variants were widely employed in time series prediction tasks\cite{elman1990finding,hochreiter1997long}. However, to address their limitations in long-sequence forecasting, several studies have adopted network architectures based on transformers and attention modules. For instance, Informer\cite{zhou2021informer} introduced the probsparse self-attention mechanism to reduce the computational complexity of traditional self-attention, thereby addressing the memory bottleneck in long sequence tasks. Autoformer\cite{wu2021autoformer} integrated the idea of time-series decomposition into the Transformer framework and proposed the Auto-Correlation mechanism to replace traditional self-attention. \\
~ \\
\textbf{Pre-training and Fine-tune for Time Series Model}.
The advancement of pretrained models in natural language processing and computer vision has significantly improved the understanding of these modalities \cite{dong2019unified,selva2023video}. Inspired by this progress, the field of time series forecasting has increasingly adopted self-supervised learning techniques \cite{zhang2024self}. Notable approaches include masked reconstruction \cite{zerveas2021transformer,nietime} and contrastive learning \cite{zhang2022self,yue2022ts2vec,yang2023dcdetector}. 
Recently, a shift towards general pretraining on extensive datasets for time series models has been observed. Current research endeavors utilizing foundational time series models for broad-spectrum predictions: Moment \cite{goswami2024moment} stands out by employing a masked reconstruction technique, pre-trained on datasets with up to 27 billion data points, through a model architecture of 385 million parameters.
Chronos \cite{ansari2024chronos} offering pre-trained models with up to 710 million parameters. Available in four sizes, these models effectively meet diverse application needs.
Models like TimesFM \cite{das2023decoder} and Lag-Llama \cite{rasul2023lag}  which based on decoder-only architecture     also achieved excellent pretraining performance. Time-MoE \cite{shi2024time} marks a breakthrough with its scalable architecture using a sparse mixture of experts, boosting efficiency for large-scale forecasts. Leveraging the Time-300B dataset, which includes 300 billion time points, it scales to 2.4 billion parameters, reducing inference costs significantly.\\
\\
\textbf{Frequency Domain Analysis for Time series Models}. A series of mathematical methods centered around Fourier analysis\cite{stein1971introduction,duoandikoetxea2024fourier} facilitate the transformation of time series data from the time domain into the frequency domain. As the field of temporal prediction continues to evolve, an increasing number of models are transitioning from learning time-domain signals to focusing on frequency-domain learning. 
For example, Zhou et al. proposed FEDformer\cite{zhou2022fedformer}, a model that integrates a frequency domain enhancement module with transformers for improved prediction. Wu et al. introduced TimesNet\cite{wu2023timesnettemporal2dvariationmodeling}, which enhances the understanding and learning of time series by decomposing them into multiple periods using Fourier Transform. Xu et al. applied a low-pass filter to eliminate high-frequency noise from the signal, followed by utilizing complex linear layers for prediction, in their model FITS\cite{xu2023fits}.

However, most current methods rely on directly converting time-series signals into the frequency domain through techniques like Fourier Transform, and then inputting these into the model for learning. This approach does not enable the model to intrinsically learn the capability of transforming signals between the time and frequency domains. Additionally, it poses the risk of spectral leakage\cite{oppenheim1999discrete,smith1997scientist} when learning from signals of fixed length. To address this issue, we believe that combining a frequency-domain expert network with a mixture of experts (MoE) architecture is an effective solution. By allowing different experts to learn the frequency domain characteristics of signals, this approach enhances the model's understanding of the intrinsic properties of the signals.


\section{method}

Given a historical input sequences \( X^i = [x^i_{1}, x^i_{2}, \cdots, x^i_{T_x}] \) of fixed length \( T_x\), we forcast the sequence values in a future window of specified length \( T_y \), represented as \( Y^i = [y^i_{1}, y^i_{2}, \cdots, y^i_{T_y}] \): 
\begin{equation}
    Y^i = \mathbf{f}(X^i;\mathbf{\theta}),
\end{equation}
where \(\mathbf{\theta}\) represents the time series model parameters. 

\begin{figure*}
    \centering
    \centerline{\includegraphics[width=0.85\linewidth]{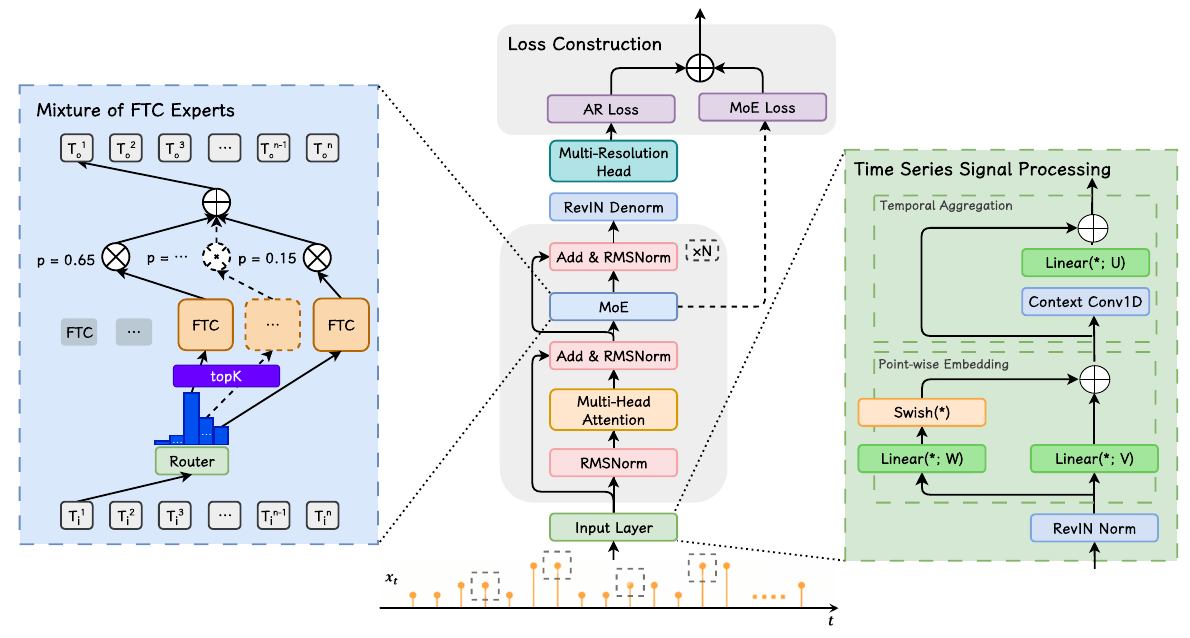}}  
    \caption{Proposed architecture. Our network is based on the successful Time-MoE\cite{shi2024time} architecture. The proposed time series signal processing and Frequency-Time Cell (FTC) for Domain Experts are contained inside the blue and green boxes. By introducing RevIN and time aggregation methods, the ability of the model to deal with non-stationarity and the applicability of the model to deal with variable-length series are improved. FTC enhancing the model's ability to capture the intrinsic properties of the signal's time and frequency domain, which contributes to improved predictive performance of the model.}
    \label{fig:framework}
\end{figure*}

We adopt the widely-regarded MoE architecture, and design expert networks utilizing Fourier analysis to capture intrinsic frequency domain characteristics as well as time domain characteristics of time series. 
An overview of our approach is given in Figure\ref{fig:framework}.

\subsection{\textbf{Time Series Signal Processing}}
For model input, we apply tokenization to the time series data. This approach allows for training standard language models on the ``language of time series'' without requiring alterations to the architecture of LLMs.  In addition, we have added RevIN and Temporal Aggregation modules.

Non-stationarity is an inherent property of real-world time series data, resulting in varying distributions even within short context windows of an individual time series. To effectively counteract the challenges posed by shifts in data distribution, we implement RevIN\cite{kim2021reversible} at the data input layer. 
Initially, the input data $X^i$ is normalized using instance-specific mean and standard deviation:
\begin{align}
    Mean[X^i] &= \frac{1}{T_x} \sum_{t=1}^{T_x} x_t^i,  \quad
    Var[X^i] = \frac{1}{T_x}\sum_{t=1}^{T_X}( x_t^i-Mean[X^i]),\\
    X^i &= \alpha (\frac{X^i-Mean[X^i]}{\sqrt{Var[X^i]}}) + \beta ,
\end{align}
where \(\alpha\) and \(\beta\) represent learnable affine parameters.
To better capture the original data distribution, a symmetric denormalization layer scales and shifts the output:

\begin{align}
    Y^i_{\textit{revin}} = \sqrt{Var[X^i+\epsilon]} \cdot (\frac{Y^i-\beta}{\alpha}) + Mean[X^i].
\end{align}

This normalization  is enhanced with learnable affine transformations,  ensures consistent mean and variance in the normalized sequence, reducing non-stationary components in input time series and enhancing the model's ability to learn intrinsic stationary signal attributes in the frequency domain.



Then, we employ point-wise embedding to represent the input floating-point time series signals. For each input sequence $X^i = [x^i_0, x^i_1, \cdots, x^i_{T_x}]$, where $x^i_{j} \in \mathbb{R}$, the embedding \( E^i = [E^i_0, E^i_1, \cdots, E^i_{w}] \in \mathbb{R}^{w\times h}  \) is generated through the following steps:  

Each scalar $x^i_{j}$ is independently mapped to an \( h \)-dimensional latent space via two parallel linear transformations:
\begin{equation}
Z_{j}^{W}=W x_{j}^{i}, \quad Z_{j}^{V}=V x_{j}^{i}, \quad W, V \in \mathbb{R}^{h \times 1}
\end{equation}
decoupling the feature subspaces while preserving computational efficiency. To dynamically balance nonlinear transformations and linear projections, we introduce a learnable mixing coefficient  $\alpha \in [0,1]$, with a trainable scalar $\beta$. The gated output is formulated as:
\begin{equation}
    E^i_j = \alpha \cdot \mathbf{swish}(Z_{j}^{W}) + (1-\alpha) \cdot Z_{j}^{V}, \quad \alpha = \mathbf{Sigmod}(\beta)
\end{equation}

This adaptive mechanism mitigates potential saturation issues in fixed multiplicative gating while retaining point-wise computation. Despite the point-wise design, temporal locality is critical for time series.  We apply dilated depthwise convolutions with kernel size \(K=3\) and dilation rate \(d=2\) across the time axis \(j\):
\begin{align}
\hat{E}_{j}^{i} &= E_{j}^{i} + \sum_{k=0}^{K-1} C_{k} \cdot E_{j-k \cdot d}^{i}, \quad C \in \mathbb{R}^{K \times h}, \\
E^{i} &= \mathbf{Linear}\left(\hat{E}^{i}\right) \in \mathbb{R}^{w \times h},
\end{align}
where  \(C\) denotes lightweight dilated convolution filters.  The dilation expands the receptive field without increasing parameters, and a residual connection that preserves the original point-wise embedding, ensures the model retains the critical local signal.
\begin{figure*}[h]
    \centering
    \centerline{\includegraphics[width=0.8\linewidth]{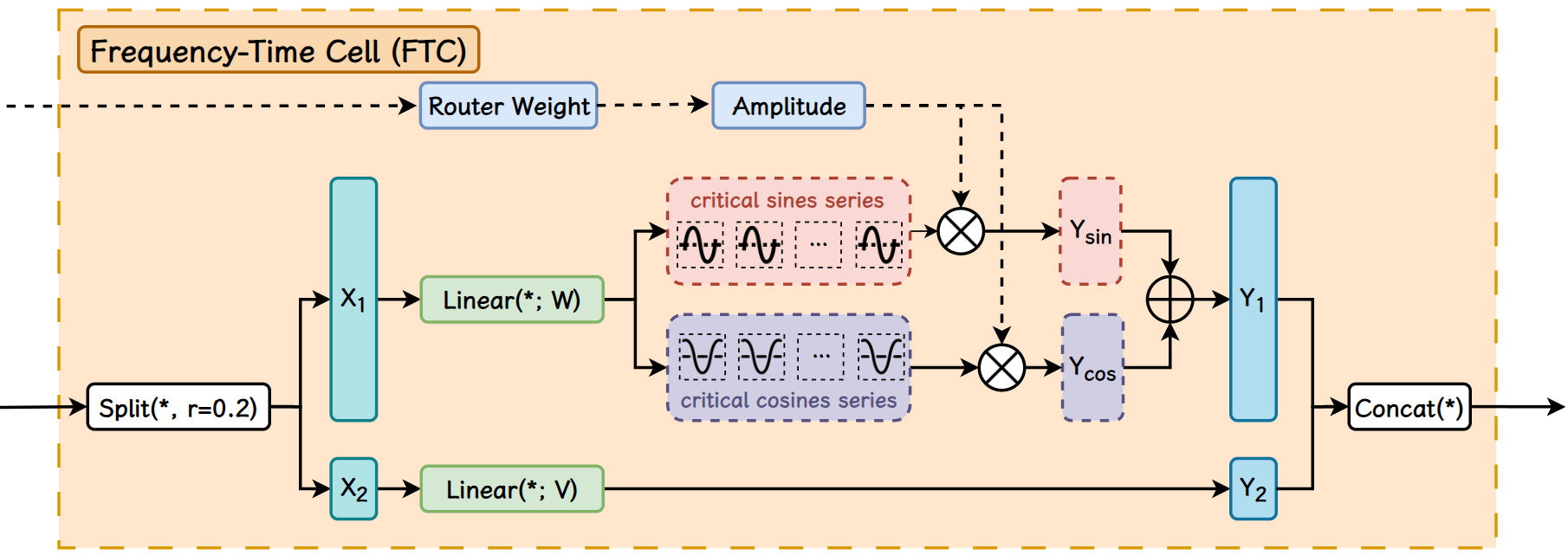}}
    \caption{Details of the FTC module. We incorporate a Frequency-Time Cell (FTC) within the MoE framework as a specialized module to enable models to simultaneously capture both frequency-domain and time-domain characteristics of signals. By mapping a segment of the input signal through neural network layers and harmonic functions into a frequency-domain space, and subsequently integrating both time-domain and frequency-domain features before feeding them into the subsequent layers, we aim to enhance the model's capacity for representation and learning.}
    \label{c}
\end{figure*}
\subsection{\textbf{Frequency-Time Cell for Domain Experts}}
We added a Frequency-Time Cell (FTC) within the MoE framework as a specialized module to enable models to simultaneously capture both frequency-domain and time-domain characteristics of signals. 

A time series within a context window can be regarded as a non-periodic discrete-time signal, whose spectrum corresponds to a periodic signal in continuous time. our network learns two key pieces of information: 

\begin{enumerate}
    \item Critical frequency components \(\{\omega_{i} | i = 1, \cdots, k\}\),
    \item Amplitudes of these critical frequency \(\{a_i | i = 1, \cdots, k\}\). 
\end{enumerate}

The original time-domain signal can then be expressed as:
\begin{equation}
    x_n = \sum_{i=1}^{k} \left(a_i e^{-j\omega_in}) \right).
\end{equation}

In practice, the spectrum of a discrete non-periodic signal is inherently continuous, making it challenging to identify the most critical spectral components. To overcome this issue, pretraining is employed to enable the model to learn \(k \times h\) distinct critical harmonic frequencies, where \(k\) is the number of experts and \(h\) is the number of distinct critical harmonic frequencies each expert holds. \par

For a given time point, the MoE routing algorithm selects the most significant \(k \times  h\) harmonics, subsequently performing a weighted sum by column, resulting in an \(h\)-dimensional vector representing the signal's periodicity. The routing weights correspond to the amplitudes of the different harmonics. Assume that the most critical frequencies learned by the \(k\) chosen experts are:
\begin{equation}
    F = [\Omega_{1}, \Omega_{2}, \cdots, \Omega_{k}],
\end{equation}
where \(\Omega_{i} = [\omega^i_1,\omega^i_2,\cdots,\omega^i_{h}]\) denotes the \(h\) key frequencies learned by the \(i\)-th expert. Letting the routing weights be \(A = [\alpha_1, \alpha_2, \cdots, \alpha_{k}]\), for the \(n\)-th token, the MoE architecture yields \(h\) harmonic combinations representing the signal, each comprising \(k\) primary harmonic frequencies:
\begin{equation}
    \underset{n = 1, 2, \cdots, h}{H[n]} = \sum_{i=1}^{k} {\alpha_i e^{-j\omega_{i}n}}.
\end{equation}

The harmonic combination vector is concatenated with the time-domain signal vector to form a representation, serving as the output of the intermediate layer. 

In the network architecture, input from the preceding layer is synchronously distributed to Frequency Time Cell Network (FTC). As shown in Figure \ref{c}, given the presence of positional encoding in the time-domain signal of each layer, inputs are initially divided into two parts: one for frequency-domain modeling and another using a linear layer for temporal feature modeling. For frequency-domain modeling, segments \(X_{t}\) are mapped to \(X_f\):
\begin{equation}
    X_f = \textit{Linear} (X_t;W),
\end{equation}
converting \(X_f\) back to the time domain through harmonic basis functions \(e^{jwn}\), the output for the \(k\)-th expert is:
\begin{equation}
    X_t^{k} = e^{jX_f^k}.
\end{equation}

Since \(X_{t}\) is a time-domain signal containing positional encoding, it effectively learns periodicity \(\omega_i\), the design's foundation. The causal multi-head attention preceding the FTC captures context information, enabling the modeling of signal periodicity.

Using the MoE routing algorithm, the model selects and weights various frequency features, imparting periodic modeling ability. To enhance the robustness of the model, we employed two of the most common harmonic basis functions, cosine and sine, and performed a straightforward summation of these functions in the implementation.  


\begin{figure*}[h!]
    \centering
    \centerline{\includegraphics[width=0.80\linewidth]
    {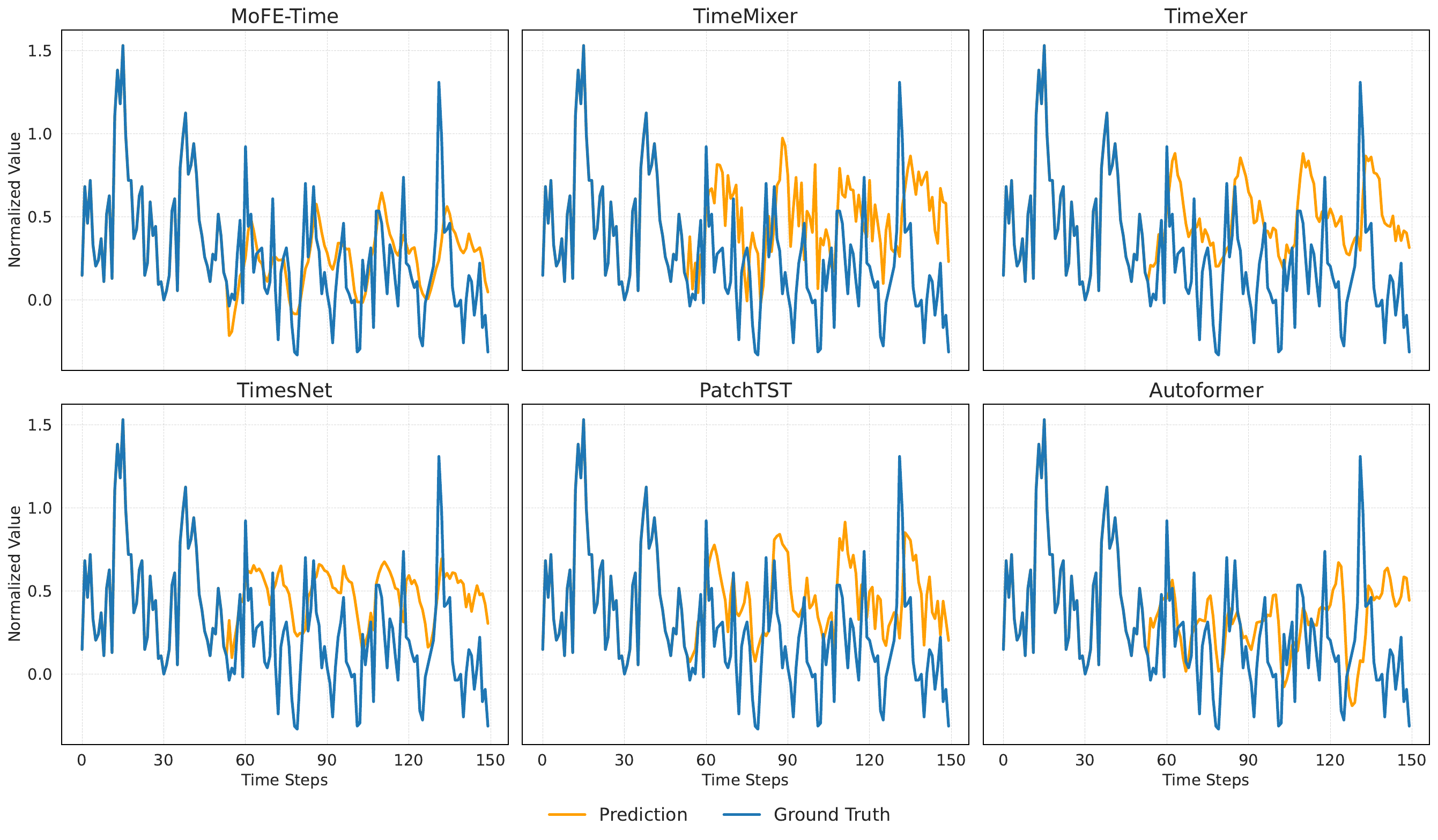}}
    \caption{Time series forecasting cases from ETTh1 by different models with context-predict length = 512-96.}
    \label{main_result_vision}
\end{figure*}
\subsection{\textbf{Supervision}}

We supervised all predictions of time series by two key components: autoregressive loss and an expert-level auxiliary loss tailored for the MoE architecture.

Autoregressive Loss:
\begin{equation}
\mathcal{L}_{\mathrm{ar}}(y_t, \hat{y}_t) =
\begin{cases} 
\frac{1}{2} (y_t - \hat{y}_t)^2, & \text{if } |y_t - \hat{y}_t| \leq \delta, \\
\delta \times (|y_t - \hat{y}_t| - \frac{1}{2} \times \delta), & \text{otherwise},
\end{cases}
\end{equation}
where \( \hat{y}_t \) denotes the actual data point at time \( t \), and \( y_t \) is the predicted value.

MoE Auxiliary Loss:

\begin{equation}
\mathcal{L}_{\text{aux}} = N \sum_{i=1}^{N} f_i P_i, \quad
f_i = \frac{1}{KT} \sum_{t=1}^{T} \mathbb{I}, \quad 
P_i = \frac{1}{T} \sum_{t=1}^{T} s_{i,t},
\end{equation}
where \( f_i \) represents the fraction of tokens dispatched to expert \( i \), and \( P_i \) is the corresponding routing probability. 
Finally, the combined loss is as follows:
\begin{equation}
    \mathcal{L} = \frac{1}{T_y} \sum_{t=1}^{T_y} \mathcal{L}_{\text{ar}} ( \mathbf{Y_{\textit{revin}, \textit{t}} }, \mathbf{\hat{Y}_\textit{t}}   )+ \alpha \mathcal{L}_{\text{aux}},
\end{equation}
where $\alpha$ is a hyperparameter that balances the contributions of temporal learning and expert-level regularization, following Time-MoE\cite{shi2024time}. 


\begin{table}[h]
    \centering

    \caption{Key Statistics of Evaluation Datasets.}
    \label{key_statistics_of_evaluation_datasets}
 
    \begin{small}
    \begin{tabular}{c|c|cccc}
    \midrule
    \textbf{Source}&\textbf{Dataset} & \textbf{Domain} & \textbf{Timesteps} & \textbf{Granularity} \\
    \midrule
     \cite{zhou2021informer}&ETTh1\&2      & Energy  & 17,420 & 1 hour\\
     \cite{zhou2021informer}&ETTm1\&2      & Energy  & 69,680 & 15 min\\
     \cite{angryk2020multivariate}&Weather       & Climate & 52,696 & 10 min\\
     \cite{exchange}&Exchange & Finance & 7,588  & 1  day\\
    \midrule
    \textbf{Ours}&NEV\_sales     & Sales  & 339,856 & 1 day\\
    \hline
    \end{tabular}
    \end{small}
\end{table}
\section{Experiments}

\subsection{Datasets and Setup}
\begin{table*}[h]
\begin{small}
\centering
\setlength{\tabcolsep}{2.4pt} 
\renewcommand{\arraystretch}{0.84} 
\setlength{\aboverulesep}{1.2pt}
\setlength{\belowrulesep}{1.2pt}
\caption{Full results of both long-term and short-term time series forcasting experiments. The lower MSE and MAE indicates a better performance. \textcolor{red}{Red}:the best, \textcolor{blue}{\underline{Blue}:} the 2th best.}
\label{mainresult}

\begin{tabular}{c|c|cc|cc|cc|cc|cc|cc|cc } 
\toprule
\multicolumn{2}{c}{\textbf{Models}} & \multicolumn{2}{c}{\textbf{MoFE-time}} & \multicolumn{2}{c}{\textbf{Time-MoE}} & \multicolumn{2}{c}{\textbf{TimeMixer}} & \multicolumn{2}{c}{\textbf{TimeXer}} & \multicolumn{2}{c}{\textbf{TimesNet}} & \multicolumn{2}{c}{\textbf{PatchTST}} & \multicolumn{2}{c}{\textbf{AutoFormer}} \\ 
\multicolumn{2}{c}{\textbf{Metrics}} & \textbf{MSE} & \textbf{MAE} & \textbf{MSE} & \textbf{MAE} & \textbf{MSE} & \textbf{MAE} & \textbf{MSE} & \textbf{MAE} & \textbf{MSE} & \textbf{MAE} & \textbf{MSE} & \textbf{MAE} & \textbf{MSE} & \textbf{MAE} \\ 

\midrule
\multirow{8}{*}{\textbf{ETTh1}} & 12 & \textcolor{red}{\textbf{0.2664}} & \textcolor{red}{\textbf{0.3241 }} & 0.2836  & 0.3507  & 0.5738  & 0.4991  & 0.2768  & 0.3357  & 0.3305  & 0.3811  & \textcolor{blue}{\underline{0.2728}}  & \textcolor{blue}{\underline{0.3325}}  & 0.3800  & 0.4267  \\ 
 & 24 & \textcolor{red}{\textbf{0.2778 }} & \textcolor{red}{\textbf{0.3334}} & 0.3315  & 0.3679  & 0.4049  & 0.4238  & 0.3063  & 0.3578  & 0.3482  & 0.3930  & \textcolor{blue}{\underline{0.3008}}  & \textcolor{blue}{\underline{0.3554}}  & 0.4007  & 0.4350  \\ 
 & 48 & \textcolor{red}{\textbf{0.3116 }} & \textcolor{red}{\textbf{0.3593 }} & \textcolor{blue}{\underline{0.3201}}  & \textcolor{blue}{\underline{0.3635}}  & 0.3595  & 0.3878  & 0.3548  & 0.3858  & 0.3792  & 0.4077  & 0.3476  & 0.3855  & 0.4337  & 0.4564  \\ 
 & 96 & \textcolor{red}{\textbf{0.3373 }} & \textcolor{red}{\textbf{0.3802 }} & \textcolor{blue}{\underline{0.3604}}  & \textcolor{blue}{\underline{0.3960}}  & 0.3698  & 0.3991  & 0.3947  & 0.4215  & 0.4425  & 0.4572  & 0.3959  & 0.4191  & 0.5599  & 0.5283  \\ 
 & 192 & \textcolor{red}{\textbf{0.3805 }} & \textcolor{red}{\textbf{0.4106 }} & \textcolor{blue}{\underline{0.3855}}  & \textcolor{blue}{\underline{0.4129}}  & 0.4481  & 0.4290  & 0.4651  & 0.4757  & 0.4915  & 0.4891  & 0.5053  & 0.4953  & 0.7237  & 0.6461  \\ 
 & 336 & \textcolor{blue}{\underline{0.4141}}  & \textcolor{blue}{\underline{0.4362}}  & \textcolor{red}{\textbf{0.4067}} & \textcolor{red}{\textbf{0.4328 }} & 0.4840  & 0.4690  & 0.6162  & 0.5705  & 0.9422  & 0.7280  & 0.6998  & 0.6114  & 0.9643  & 0.7689  \\ 
 & 720 & \textcolor{red}{\textbf{0.4531 }} & \textcolor{red}{\textbf{0.4655 }} & \textcolor{blue}{\underline{0.4568}}  & \textcolor{blue}{\underline{0.4762}}  & 0.4980  & 0.5000  & 1.0342  & 0.7512  & 1.1941  & 0.8627  & 1.1119  & 0.7965  & 1.4022  & 0.9922  \\ 
 \rowcolor{gray!20}   & \textbf{AVG} & \textcolor{red}{\textbf{0.3487 }} & \textcolor{red}{\textbf{0.3870 }} & \textcolor{blue}{\underline{0.3635}}  & \textcolor{blue}{\underline{0.4000}}  & 0.4483  & 0.4440  & 0.4926  & 0.4712  & 0.5897  & 0.5313  & 0.5192  & 0.4851  & 0.6949  & 0.6077  \\ 
 \midrule

 \multirow{7}{*}{\textbf{ETTh2}}& 12 & 0.1407  & 0.2425  & 0.1640  & 0.2584  & 0.1537  & 0.2558  & \textcolor{red}{\textbf{0.1332 }} & \textcolor{blue}{\underline{0.2326}}  & 0.1516  & 0.2546  & \textcolor{blue}{\underline{0.1351}}  & \textcolor{red}{\textbf{0.2319} } & 0.2182  & 0.3135  \\ 
 & 24 & \textcolor{red}{\textbf{0.1717 }} & \textcolor{blue}{\underline{0.2646}}  & 0.2369  & 0.3015  & 0.1763  & 0.2693  & \textcolor{blue}{\underline{0.1721}}  & \textcolor{red}{\textbf{0.2632 }} & 0.2090  & 0.2963  & 0.1754  & 0.2673  & 0.2911  & 0.3749  \\ 
 & 48 & \textcolor{blue}{\underline{0.2375}}  & \textcolor{red}{\textbf{0.3064 }} & 0.2609  & 0.3172  & 0.2589  & 0.3249  & \textcolor{red}{\textbf{0.2288 }} & \textcolor{blue}{\underline{0.3107}}  & 0.2795  & 0.3452  & 0.2448  & 0.3197  & 0.3805  & 0.4342  \\ 
 & 96 & 0.3073  & \textcolor{blue}{\underline{0.3523}}  & 0.3516  & 0.3880  & \textcolor{red}{\textbf{0.2800 }} & \textcolor{red}{\textbf{0.3506 }} & \textcolor{blue}{\underline{0.3056}}  & 0.3641  & 0.3840  & 0.4206  & 0.3063  & 0.3617  & 0.4964  & 0.5134  \\ 
 & 192 & \textcolor{red}{\textbf{0.3890 }} & \textcolor{blue}{\underline{0.4181}}  & 0.4249  & 0.4335  & 0.4454  & 0.4540  & \textcolor{blue}{\underline{0.3938}}  & \textcolor{red}{\textbf{0.4141 }} & 0.4146  & 0.4365  & 0.3987  & 0.4228  & 1.5797  & 0.9588  \\ 
 & 336 & 0.5140  & \textcolor{red}{\textbf{0.4798 }} & 0.5256  & \textcolor{blue}{\underline{0.4854}}  & \textcolor{red}{\textbf{0.4878 }} & 0.4951  & \textcolor{blue}{\underline{0.5270}}  & 0.5235  & 0.6521  & 0.5902  & 0.5615  & 0.5293  & 1.3183  & 0.9030  \\ 
 & 720 & \textcolor{red}{\textbf{0.5428 }} & \textcolor{red}{\textbf{0.5050 }} & \textcolor{blue}{\underline{0.5851}}  & \textcolor{blue}{\underline{0.5257}}  & 1.7088  & 0.8594  & 1.6449  & 0.9109  & 0.8547  & 0.7072  & 1.2430  & 0.7891  & 0.9152  & 0.7564  \\ 
 \rowcolor{gray!20}    & \textbf{AVG} & \textcolor{red}{\textbf{0.3290 }} & \textcolor{red}{\textbf{0.3669 }} & \textcolor{blue}{\underline{0.3641}}  & \textcolor{blue}{\underline{0.3871}}  & 0.5016  & 0.4299  & 0.4865  & 0.4313  & 0.4208  & 0.4358  & 0.4378  & 0.4174  & 0.7428  & 0.6077  \\ 
  \midrule
\multirow{8}{*}{\textbf{ETTm1}} & 12 & 0.2130  & 0.2801  & 0.2174  & 0.2812  & 0.2313  & 0.3003  & \textcolor{red}{\textbf{0.1808 }} & \textcolor{red}{\textbf{0.2601 }} & \textcolor{blue}{\underline{0.1979}}  & \textcolor{blue}{\underline{0.2758}}  & 0.2296  & 0.2965  & 0.3898  & 0.4092  \\ 
 & 24 & 0.2197  & 0.2929  & 0.2653  & 0.3219  & 0.2203  & 0.2964  & \textcolor{red}{\textbf{0.2092 }} & \textcolor{red}{\textbf{0.2806 }} & 0.2354  & 0.3066  & \textcolor{blue}{\underline{0.2151}}  & \textcolor{blue}{\underline{0.2897}}  & 0.4814  & 0.4520  \\ 
 & 48 & \textcolor{red}{\textbf{0.2536 }} & 0.3247  & 0.3004  & 0.3439  & 0.2598  & \textcolor{blue}{\underline{0.3228}}  & 0.2695  & 0.3284  & 0.2849  & 0.3445  & \textcolor{blue}{\underline{0.2594}}  & \textcolor{red}{\textbf{0.3214 }} & 0.5144  & 0.4969  \\ 
 & 96 & \textcolor{red}{\textbf{0.2938 }} & \textcolor{blue}{\underline{0.3516}}  & 0.3192  & 0.3729  & 0.2981  & 0.3560  & 0.3349  & 0.3728  & 0.3504  & 0.3824  & \textcolor{blue}{\underline{0.2943}}  & \textcolor{red}{\textbf{0.3485 }} & 0.5906  & 0.5333  \\ 
 & 192 & \textcolor{red}{\textbf{0.3326 }} & \textcolor{red}{\textbf{0.3809 }} & \textcolor{blue}{\underline{0.3585}}  & 0.4007  & 0.3594  & \textcolor{blue}{\underline{0.3918}}  & 0.3746  & 0.4088  & 0.5194  & 0.4757  & 0.3920  & 0.4139  & 0.6939  & 0.5735  \\ 
 & 336 & \textcolor{red}{\textbf{0.4001 }} & \textcolor{red}{\textbf{0.4332 }} & \textcolor{blue}{\underline{0.4038}}  & \textcolor{blue}{\underline{0.4330}}  & 0.4570  & 0.4613  & 0.4994  & 0.4635  & 0.6174  & 0.5139  & 0.4749  & 0.4758  & 0.6569  & 0.5765  \\ 
 & 720 & \textcolor{blue}{\underline{0.5358}}  & 0.5142  & 0.5453  & \textcolor{blue}{\underline{0.5009}}  & \textcolor{red}{\textbf{0.5022 }} & \textcolor{red}{\textbf{0.4824 }} & 0.5696  & 0.5082  & 0.8081  & 0.6228  & 0.6085  & 0.5499  & 0.8389  & 0.6787  \\ 
 \rowcolor{gray!20}    & \textbf{AVG} & \textcolor{red}{\textbf{0.3212 }} & \textcolor{red}{\textbf{0.3682 }} & 0.3443  & 0.3792  & \textcolor{blue}{\underline{0.3326}}  & \textcolor{blue}{\underline{0.3730}}  & 0.3483  & 0.3746  & 0.4305  & 0.4174  & 0.3534  & 0.3851  & 0.5951  & 0.5314  \\ 
 \midrule
\multirow{8}{*}{\textbf{ETTm2}} & 12 & 0.0873  & 0.1796  & 0.1007  & 0.1996  & 0.0881  & 0.1834  & \textcolor{red}{\textbf{0.0766 }} & \textcolor{red}{\textbf{0.1659 }} & 0.0826  & 0.1763  & \textcolor{blue}{\underline{0.0817}}  & \textcolor{blue}{\underline{0.1724}}  & 0.1215  & 0.2374  \\ 
 & 24 & 0.1101  & 0.2153  & 0.1613  & 0.2508  & 0.1042  & 0.2013  & \textcolor{red}{\textbf{0.0956 }} & \textcolor{red}{\textbf{0.1905 }} & 0.1063  & 0.2049  & \textcolor{blue}{\underline{0.1019}}  & \textcolor{blue}{\underline{0.1984}}  & 0.1805  & 0.2833  \\ 
 & 48 & 0.1387  & 0.2399  & 0.1574  & 0.2534  & \textcolor{blue}{\underline{0.1305}}  & \textcolor{blue}{\underline{0.2281}}  & \textcolor{red}{\textbf{0.1286 }} & \textcolor{red}{\textbf{0.2234} } & 0.1459  & 0.2424  & 0.1329  & 0.2315  & 0.2427  & 0.3271  \\ 
 & 96 & 0.1887  & 0.2780  & 0.2577  & 0.3196  & \textcolor{red}{\textbf{0.1656 }} & \textcolor{red}{\textbf{0.2574 }} & 0.1801  & \textcolor{blue}{\underline{0.2624}}  & 0.1912  & 0.2765  & \textcolor{blue}{\underline{0.1728}}  & 0.2652  & 0.2863  & 0.3655  \\ 
 & 192 & \textcolor{blue}{\underline{0.2491}}  & 0.3274  & 0.2696  & 0.3377  & \textcolor{red}{\textbf{0.2376 }} & \textcolor{red}{\textbf{0.3114 }} & 0.2557  & \textcolor{blue}{\underline{0.3213}}  & 0.2856  & 0.3494  & 0.2664  & 0.3291  & 0.4641  & 0.4863  \\ 
 & 336 & \textcolor{red}{\textbf{0.2944 }} & \textcolor{blue}{\underline{0.3560}}  & 0.3653  & 0.4046  & \textcolor{blue}{\underline{0.2966}}  & \textcolor{red}{\textbf{0.3505 }} & 0.3000  & 0.3599  & 0.3678  & 0.4185  & 0.3704  & 0.3909  & 0.6277  & 0.6060  \\ 
 & 720 & \textcolor{red}{\textbf{0.3810 }} & \textcolor{blue}{\underline{0.4253}}  & 0.4030  & 0.4450  & \textcolor{blue}{\underline{0.3915}}  & \textcolor{red}{\textbf{0.4090 }} & 0.4167  & 0.4266  & 0.6244  & 0.5227  & 0.5001  & 0.4748  & 1.0908  & 0.8019  \\ 
 \rowcolor{gray!20}    & \textbf{AVG} & \textcolor{blue}{\underline{0.2070}}  & 0.2888  & 0.2450  & 0.3158  & \textcolor{red}{\textbf{0.2020 }} & \textcolor{red}{\textbf{0.2773 }} & 0.2076  & \textcolor{blue}{\underline{0.2786}}  & 0.2577  & 0.3129  & 0.2323  & 0.2946  & 0.4305  & 0.4439  \\ 
 \midrule
\multirow{8}{*}{\textbf{Weather}} & 12 & \textcolor{blue}{\underline{0.0723}}  & \textcolor{blue}{\underline{0.0977}}  & 0.0880  & 0.1117  & 0.0853  & 0.1144  & \textcolor{red}{\textbf{0.0715 }} &\textcolor{red}{ \textbf{0.0928 }} & 0.0873  & 0.1254  & 0.0780  & 0.1003  & 0.1398  & 0.2172  \\ 
 & 24 & \textcolor{blue}{\underline{0.0924}}  & \textcolor{blue}{\underline{0.1248}}  & 0.1285  & 0.1583  & 0.1120  & 0.1569  & \textcolor{red}{\textbf{0.0923 }} & \textcolor{red}{\textbf{0.1241 }} & 0.1007  & 0.1454  & 0.1003  & 0.1372  & 0.1735  & 0.2605  \\ 
 & 48 & \textcolor{red}{\textbf{0.1132 }} & 0.1574  & 0.1415  & 0.1698  & 0.1254  & 0.1746  & \textcolor{blue}{\underline{0.1142}}  & \textcolor{red}{\textbf{0.1553 }} & 0.1292  & 0.1802  & 0.1216  & 0.1621  & 0.2647  & 0.3483  \\ 
 & 96 & \textcolor{blue}{\underline{0.1489}}  & \textcolor{blue}{\underline{0.1984}}  & 0.1789  & 0.2169  & \textcolor{red}{\textbf{0.1462 }} & \textcolor{red}{\textbf{0.1981 }} & 0.1520  & 0.2036  & 0.1585  & 0.2141  & 0.1561  & 0.2111  & 0.3622  & 0.4033  \\ 
 & 192 & \textcolor{red}{\textbf{0.1888 }} & \textcolor{red}{\textbf{0.2439 }} & 0.2110  & 0.2604  & 0.2073  & 0.2594  & \textcolor{blue}{\underline{0.2041}}  & \textcolor{blue}{\underline{0.2568}}  & 0.2362  & 0.2843  & 0.2330  & 0.2823  & 0.4145  & 0.4190  \\ 
 & 336 & \textcolor{red}{\textbf{0.2467 }} & \textcolor{red}{\textbf{0.2879 }} & \textcolor{blue}{\underline{0.2495}}  & \textcolor{blue}{\underline{0.2973}}  & 0.2633  & 0.3072  & 0.2743  & 0.3156  & 0.3028  & 0.3332  & 0.2960  & 0.3285  & 0.4255  & 0.4388  \\ 
 & 720 & \textcolor{red}{\textbf{0.3285 }} & \textcolor{red}{\textbf{0.3539 }} & 0.3388  & 0.3638  & \textcolor{blue}{\underline{0.3300}}  & \textcolor{blue}{\underline{0.3556}}  & 0.3533  & 0.3701  & 0.4254  & 0.4203  & 0.3780  & 0.3872  & 0.4168  & 0.4377  \\ 
  \rowcolor{gray!20}   & \textbf{AVG} & \textcolor{red}{\textbf{0.1701 }} & \textcolor{red}{\textbf{0.2092 }} & 0.1909  & 0.2254  & 0.1814  & 0.2237  & \textcolor{blue}{\underline{0.1802}}  & \textcolor{blue}{\underline{0.2169}}  & 0.2057  & 0.2433  & 0.1947  & 0.2298  & 0.3138  & 0.3607  \\ 
 \midrule
\multirow{8}{*}{\textbf{Exchange}} & 12 & 0.0159  & \textcolor{red}{\textbf{0.0106 }} & 0.0459  & 0.1399  & 0.0473  & 0.1494  & \textcolor{red}{\textbf{0.0135 }} & 0.0750  & 0.0181  & 0.0919  & \textcolor{blue}{\underline{0.0135}}  & \textcolor{blue}{\underline{0.0470}}  & 0.0294  & 0.1214  \\ 
 & 24 & 0.0432  & 0.1412  & 0.0788  & 0.1862  & 0.0444  & 0.1538  & \textcolor{blue}{\underline{0.0249}}  & \textcolor{blue}{\underline{0.1059}}  & 0.0357  & 0.1341  & \textcolor{red}{\textbf{0.0245 }} & \textcolor{red}{\textbf{0.1044 }} & 0.0532  & 0.1693  \\ 
 & 48 & 0.0804  & 0.1961  & 0.1094  & 0.2277  & \textcolor{blue}{\underline{0.0628}}  & 0.1811  & 0.0464  & \textcolor{red}{\textbf{0.1503 }} & 0.0777  & 0.2070  & \textcolor{red}{\textbf{0.0470 }} & \textcolor{blue}{\underline{0.1521}}  & 0.2422  & 0.3778  \\ 
 & 96 & \textcolor{red}{\textbf{0.0829 }} & \textcolor{red}{\textbf{0.2088 }} & 0.2007  & 0.3168  & 0.1143  & 0.2456  & 0.1068  & 0.2285  & 0.2208  & 0.3521  & \textcolor{blue}{\underline{0.0978}}  & \textcolor{blue}{\underline{0.2243}}  & 0.9976  & 0.7922  \\ 
 & 192 & \textcolor{red}{\textbf{0.1688 }} & \textcolor{red}{\textbf{0.3060 }} & 0.3140  & 0.4082  & 0.2563  & 0.3750  & 0.2212  & 0.3439  & 0.8259  & 0.6261  & \textcolor{blue}{\underline{0.2153}}  & \textcolor{blue}{\underline{0.3395}}  & 2.5804  & 1.2680  \\ 
 & 336 & \textcolor{blue}{\underline{0.5186}}  & 0.5576  & \textcolor{red}{\textbf{0.4323 }} & \textcolor{red}{\textbf{0.4908 }} & 0.5392  & 0.5573  & 0.5502  & \textcolor{blue}{\underline{0.5113}}  & 1.3678  & 0.8242  & 0.6636  & 0.6109  & 1.9066  & 1.0931  \\ 
 & 720 & \textcolor{blue}{\underline{1.0274}}  & \textcolor{blue}{\underline{0.7886}}  & \textcolor{red}{\textbf{0.7001 }} & \textcolor{red}{\textbf{0.6939 }} & 2.4504  & 1.2700  & 2.5958  & 1.2798  & 3.5770  & 1.4513  & 1.4083  & 0.9432  & 3.4996  & 1.3935  \\ 
 \rowcolor{gray!20}    & \textbf{AVG} & \textcolor{blue}{\underline{0.2767}}  & \textcolor{red}{\textbf{0.3155}} & \textcolor{red}{\textbf{0.2687 }} & 0.3519  & 0.5021  & 0.4189  & 0.5084  & 0.3849  & 0.8747  & 0.5267  & 0.3529  & \textcolor{blue}{\underline{0.3459}}  & 1.3299  & 0.7450  \\ 
 \midrule
\rowcolor{red!20}     \multicolumn{2}{c}{\textbf{Average}} & \textcolor{red}{\textbf{0.2755 }} & \textcolor{red}{\textbf{0.3226}} & \textcolor{blue}{\underline{0.2961}}  & \textcolor{blue}{\underline{0.3433}}  & 0.3613  & 0.3611  & 0.3706  & 0.3596  & 0.4632  & 0.4112  & 0.3484  & 0.3597  & 0.6845  & 0.5494  \\ 

\midrule
\rowcolor{orange!20}    \multicolumn{2}{c}{\textbf{1th Count}} & \multicolumn{2}{c}{\textcolor{red}{\textbf{39}} } & \multicolumn{2}{c}{6 } & \multicolumn{2}{c}{12 } & \multicolumn{2}{c}{\textcolor{blue}{\underline{21}} } & \multicolumn{2}{c}{0 } & \multicolumn{2}{c}{6 } & \multicolumn{2}{c}{0 } \\ 
\bottomrule
\end{tabular}
\end{small}
\end{table*}

\textbf{Pretraining Dataset.}
We select Time-300B\cite{shi2024time}, currently the largest and highest quality publicly available time series dataset. This dataset is comprised of multiple public datasets across various domains, including energy, retail, healthcare, weather, finance, transportation, and networking, as well as a portion of synthetic data. The Time-300B dataset spans a wide range of sampling frequencies, from intervals of a few seconds to up to one year. 

\textbf{Finetuing and Evaluation Dataset.}
We primarily utilized six public datasets and one proprietary dataset to evaluate the fine tune performance of the MoFE-Time. For the  public datasets, we choose ETTh1, ETTh2, ETTm1, ETTm2, Weather and Exchange rate. The ETT (Electricity Transformer Temperature) series consists of two hourly datasets (ETTh1\&h2) and two 15-minute level dataset (ETTm1\&m2). 
\par

\textbf{Commercial sector store traffic forecasting dataset.} To verify the effectiveness of MoFE-Time in real-world applications, we collected and organized daily store traffic data from the sales centers of a new energy vehicle company to construct the NEV-sales (New Energy Vehicle sales) dataset. Store traffic flow is a typical source of time series data with seasonal and cyclical patterns. We utilized traffic data from over 400 sales centers, covering the period from their inception until December 31, 2024. During the data cleaning process, we removed store data with more than 30 consecutive days of zero values and excluded non-continuous sequences where daily data completeness was below 95\%. This process resulted in the selection of 498 continuous time series at a daily granularity. The final dataset comprises a total of 330,000 time points, spanning 32 provincial-level administrative regions in China and covering major first- and second-tier cities across the country.Detailed parameters of the evaluation datasets are presented in Table \ref{key_statistics_of_evaluation_datasets}.

\textbf{Implementation details.}
The training of the MoFE-Time model primarily involves two phases: pre-training and transfer learning. First, we use the Time-300B dataset for pretraining.
We use AdamW optimizer with the learning rate=$1e-3$, weight\_decay = $0.1$, $\beta_1$ = $0.9$, $\beta_2$ = $0.95$. The epoch was set to $1$, incorporating linear learning rate warm up for $10\%$ during the training. 
The pretraining batch size was configured at $2$, utilizing $\text{bf}16$ for model training data precision.  After pre-training the model, we attempted transfer learning training on public datasets. For each transfer learning set, we conducted training for one epoch with a learning rate of $5e-6$, and eliminated the warm up process. All other parameter settings were kept consistent with the pre-training phase.

\begin{table*}[ht]
\centering
\caption{Full results of  time series forcasting experiments for NEV-sales datasets. The lower MSE and MAE indicates a better performance. \textbf{\textcolor{red}{Red}}: the best, \textcolor{blue}{\underline{Blue}:} the 2th best.}
\label{Nevs-sales}

\begin{small}
\begin{tabular}{c|cc|cc|cc|cc|cc|cc|cc } 
\hline

\textbf{Models} & \multicolumn{2}{c}{\textbf{MoFE-Time}} & \multicolumn{2}{c}{\textbf{Time-MoE}} & \multicolumn{2}{c}{\textbf{TimeMixer}} & \multicolumn{2}{c}{\textbf{TimeXer}} & \multicolumn{2}{c}{\textbf{TimesNet}} & \multicolumn{2}{c}{\textbf{PatchTST}} & \multicolumn{2}{c}{\textbf{AutoFormer}} \\
\hline
\textbf{Metrics} & \textbf{MSE} & \textbf{MAE} & \textbf{MSE} & \textbf{MAE} & \textbf{MSE} & \textbf{MAE} & \textbf{MSE} & \textbf{MAE} & \textbf{MSE} & \textbf{MAE} & \textbf{MSE} & \textbf{MAE} & \textbf{MSE} & \textbf{MAE} \\ 

    \hline
    8  & 0.2548 & 0.3822 & 0.2843 & 0.3917 & \textcolor{blue}{\underline{ 0.2094}} & \textcolor{blue}{\underline{0.3394}} & 0.2381 & 0.3603 & 0.3235 & 0.3924 & \textcolor{red}{\textbf{0.1930}} & \textcolor{red}{\textbf{0.3249}} & 0.2252 & 0.3546 \\
    \hline
    16  & \textcolor{red}{\textbf{0.1877}} &  \textcolor{red}{\textbf{0.3262}} & \textcolor{blue}{\underline{0.2234}} & \textcolor{blue}{\underline{0.3432}}  &0.2680 & 0.4051 & 0.3217 & 0.4286 & 0.4139 & 0.4411 & 0.3678 & 0.4792 & 0.2966 & 0.4197 \\
    \hline
    24 & \textbf{\textcolor{red}{0.1688}} & \textbf{\textcolor{red}{0.2995}} & \textcolor{blue}{\underline{0.2138}} & \textcolor{blue}{\underline{0.3536}}  & 0.2521 & 0.3787 & 0.2303 & 0.3549 & 0.2626 & 0.3680 & 0.2976 & 0.4325 & 0.2554 & 0.3780\\
    \hline
    
 \rowcolor{gray!20}    Average & \textbf{\textcolor{red}{0.1956}} & \textbf{\textcolor{red}{0.3284}} & \textcolor{blue}{\underline{0.2405}} & \textcolor{blue}{\underline{0.3628}}&  0.2432 & 0.3744 & 0.2634 & 0.3813 & 0.3333 & 0.4005 & 0.2862 & 0.4122 & 0.2591 & 0.3841 \\
    \hline
   
    \end{tabular}%
    \end{small}
\end{table*}

\begin{figure*}[h]
    \centering
    \centerline{\includegraphics[width=0.75\linewidth]{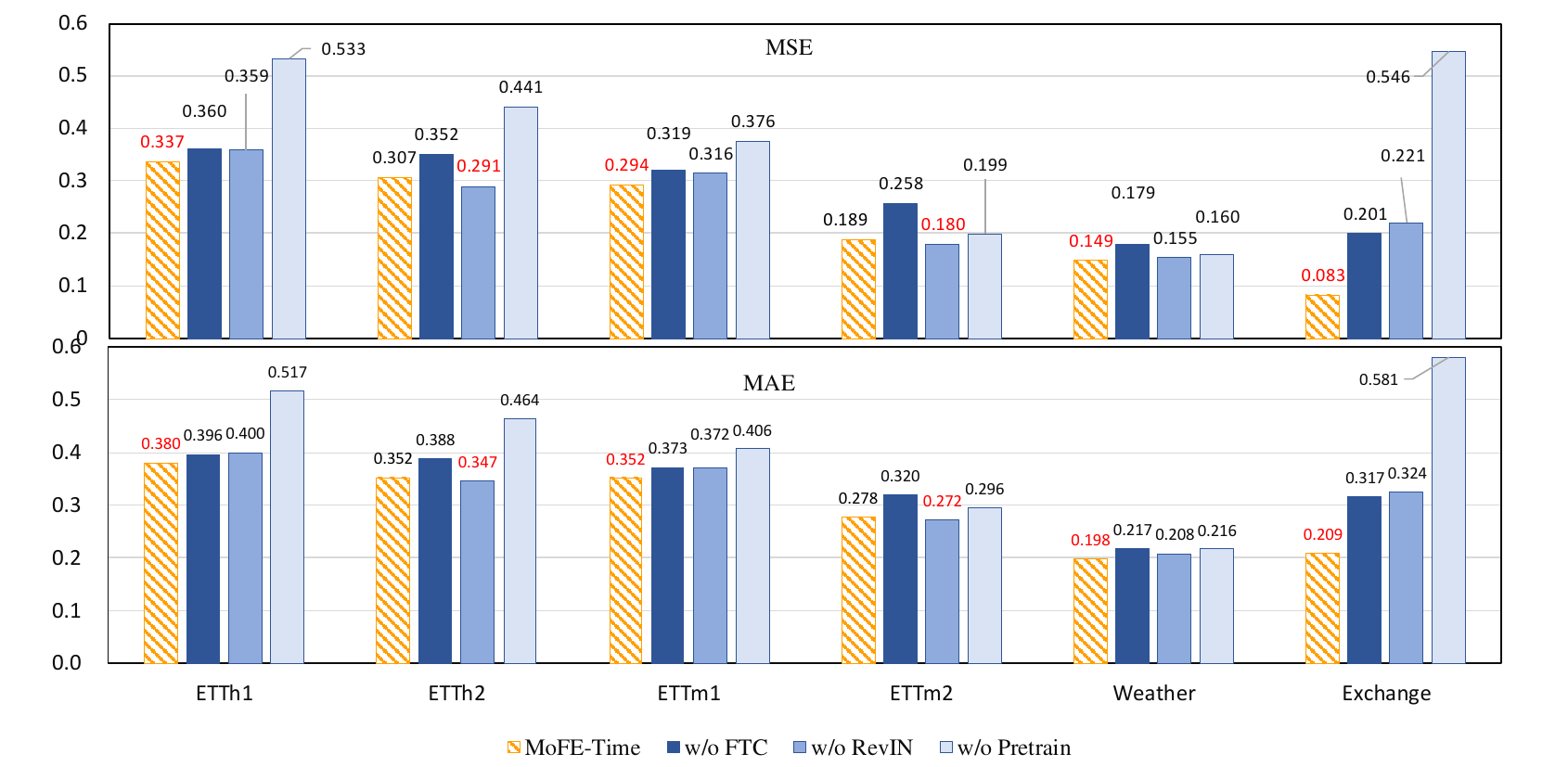}}
    \caption{The ablation study of pretrain , RevIN and FTC in MoFE-Time.}
    \label{Ablation}
\end{figure*}


\subsection{Comparison on Benchmarks}

To ensure a fair comparison, we reproduce six well-known time series forecasting models: Time-MoE \cite{shi2024time}, TimeMixer \cite{wang2024timemixer++}, TimeXer \cite{wang2024timexer}, TimesNet \cite{wu2023timesnettemporal2dvariationmodeling}, and AutoFormer \cite{wu2021autoformer}. These methods encompass both models employing a pre-training and fine-tuning architecture and networks that learn using frequency domain features.

In our evaluation setup, prediction lengths varied among {12, 24, 48, 96, 192, 336, 720}, corresponding to context lengths of {64, 128, 256, 512, 1024, 2048, 3072}. Notably, most existing studies focus on prediction lengths of 96 or larger, with corresponding context lengths of at least 256. We introduce additional evaluations with three shorter prediction lengths to provide a more comprehensive analysis. As illustrated in Figure \ref{main_result_vision}, except for the ETTm2 dataset, MoFE-Time outperforms its competitors. On average, MoFE-Time achieves MSE and MAE of \textbf{0.2755} and \textbf{0.3226}, reflecting reductions of \textbf{6.95\%} and \textbf{6.02\%} against the next best model, Time-MoE.

\subsection{Commercial Store Traffic Forecast}

In the realm of commercial forecasting, predicting store traffic plays a pivotal role and can extend to various analogous business metrics, such as online visitor counts. We specifically applied MoFE-Time to the task of forecasting foot traffic in new energy vehicle (NEV) stores, drawing on its potential to improve operational and strategic business decisions. Store traffic data often exhibits strong seasonality, long cycles, and a high level of complexity due to pervasive noise, thus posing  challenges for accurate prediction.

For the NEV-sales dataset, we conducted evaluations using prediction lengths of $8, 16, 24$, with corresponding context lengths of $56, 112, 180$.  Despite the challenges inherent in commercial datasets, MoFE-Time consistently delivers superior performance, with average MSE and MAE values recorded at $0.1956$ and $0.3284$, respectively, across the dataset, as depicted in Table \ref{Nevs-sales}. These findings underscore the model's robustness and applicability in real-world commercial environments. Our evaluation  emphasize the model's capacity to accurately model patterns in commercial datasets, thus providing a valuable tool for business intelligence and operational decision-making.



\begin{figure*}[ht]
    \centering
    \centerline{\includegraphics[width=1.00\linewidth]{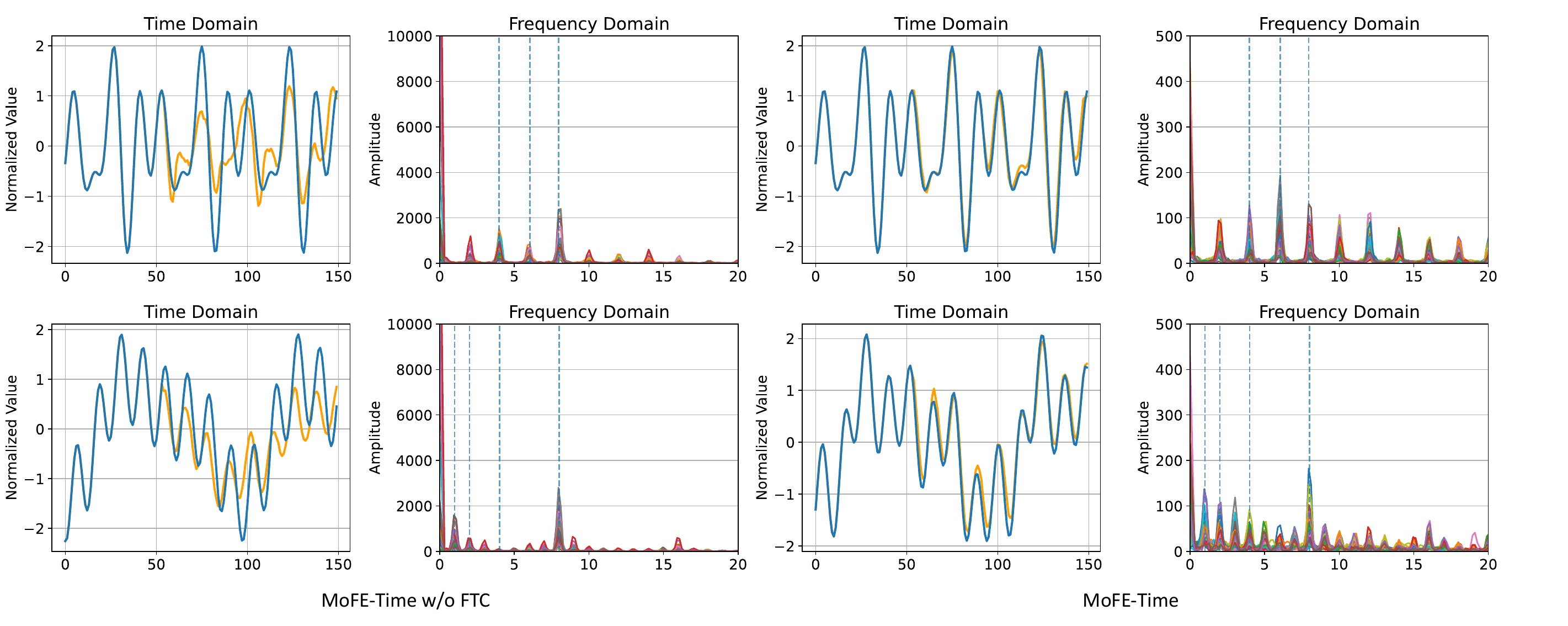}}
    \caption{The visualization of frequency features between MoFE-Time and MoFE-Time w/o FTC. In the first column, we synthesized a composite sine single with 4, 6, 8Hz, while  the second one is a composite sine single with 1, 2, 4, 8Hz.
     The figure illustrates the predictive results and spectral analysis of the MoFE-Time model and the MoFE-Time model w/o FTC for the two aforementioned harmonics at a context-predict length of 512-96. In the frequency domain plot, the \textcolor[rgb]{0.44,0.615,0.776}{blue dash} represents the ground truth frequency distribution of the harmonic.}
    \label{distribution_l_g}
\end{figure*}
\subsection{Ablations}

In this section, the ablation experiments are conducted, aiming at verifying the efficacy of the pre-training-finetuning paradigm, RevIN and frequency domain expert. 

As shown in Figure \ref{Ablation}, we performed experiments under a prediction length setting of 96 with context length of 512 on six public datasets.  From the result, we can conclude that most of those technical points contribute to improving the results. However, different technical points have different contributions:
(1) Both metrics (MES and MAE) on  all the 6 datasets get a better result if FTC was introduced in. 
(2) The introduction of RevIN  haves a positive contribution in 4 out of the 6 datasets.
(3) The pretraining stage offers the largest contribution to both MSE and MAE of the 6 datasets.
As previously mentioned, pretraining intruduce external knowledge and  FTC intruduce periodicity modeling mechanism.  Since both of the above two points are essential for time series prediction, they maintain consistent performance more consistent performance improvement acrosss different scenarios.
Since samples of Exchange-rate dataset are more unstable signals, ReVIN which was designed for mitigate alleviates signal stability issues, provides the largest contribution on exchange dataset.  Different from many other researches who primarily focus on periodic modeling,  our experiments demonstrate that external knowledge learning may be more important than  periodicity modeling, and that is why we reemphasized the pretraining-finetuning paradigm in this field.

\subsection{Visualization of the Expert Spectral Analysis}

To visualize the frequency-dimensional signal characteristics, we fetched the last layer's hidden state of frequency domain experts block on MoFE-Time of two composite signals and the feedforward layer on MoFE-Time without FTC.
We then applied the Fast Fourier Transform along the time dimension and compared the energy distribution across the frequency spectrum with the ground truth in Figure \ref{distribution_l_g}. 

Two composite sinusoidal signals are synthesized at 4, 6, 8 Hz and 1, 2, 4, 8 Hz.
The results show that the MoFE-Time model with the FTC module has superior prediction performance. The spectral characteristic energy of its FTC module is predominantly concentrated at the correct harmonic frequencies. In contrast, the traditional MoE architecture, which replaces the FTC module with a feedforward layer, demonstrates weaker predictive performance and less precise spectral distribution. This further confirms the effectiveness of incorporating frequency domain expert networks within the MoFE-Time model.

\subsection{Inference Timing}
\begin{figure}[h]
  \centering
  \includegraphics[width=1.01\linewidth]{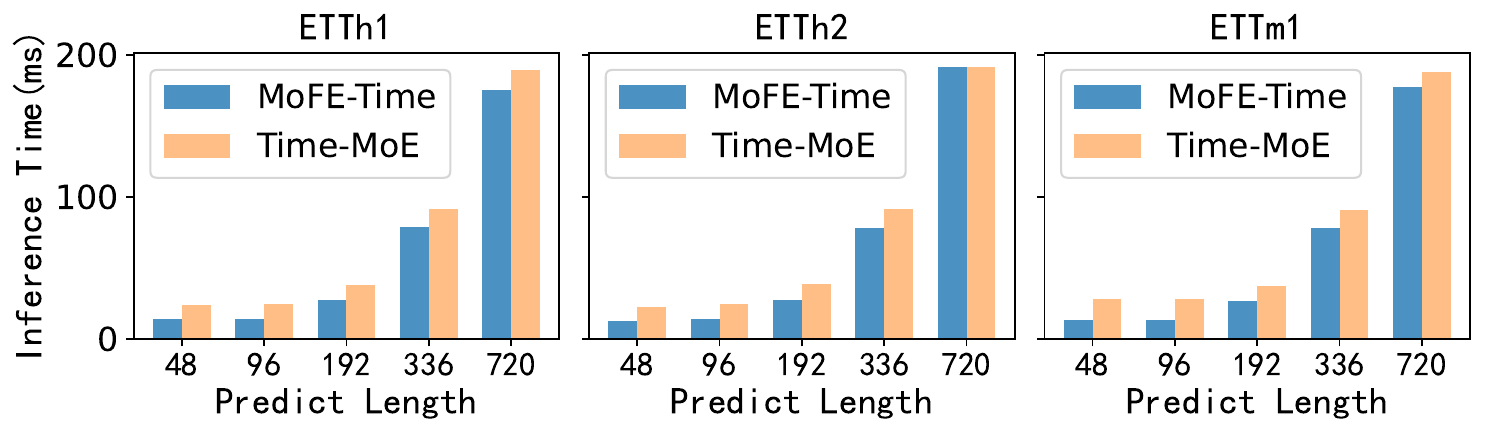}
  \caption{Comparison of Inference Speed per Data Point between MoFE-Time and TimeMoE}
  \label{time}
\end{figure}

We utilized MoFe-Time with 117.95 M parameters and Time-MoE with 113.35 M parameters (Aligned with the model version in Table \ref{mainresult}). Inference time is shown in Figure \ref{time}. In these plots, we report the inference time on the ETTh1, ETTh2, and ETTm1 datasets, we time our method and Time—MoE using a A100(40GB) GPU. MoFe-Time demonstrates equal or even faster inference speeds compared to Time-MoE under the same scale of parameters.






\section{Conclusions}
We present MoFE-Time, a novel time series forecasting model that leverages frequency domain analysis networks alongside a MoE architecture. By integrating a frequency domain analysis module within the expert networks and combining it with temporal features of signals, the model enhances its ability to learn the intrinsic properties of time series. The training architecture employs a pre-training phase followed by fine-tuning, supplemented by the integration of the RevIN method, which collectively enhances the model's predictive capabilities. 
MoFE-Time has achieved new state-of-the-art performance on a few benchmarks, and is especially robust on a proprietary dataset focusing on NEV sales, substantiating its effectiveness in real-world commercial applications forecasting scenarios, thus providing a valuable tool for business intelligence and operational decision-making.



\bibliographystyle{ACM-Reference-Format}
\bibliography{sample-base}

\end{document}